# UrbanSense

## A Framework for Quantitative Analysis of Urban Streetscapes leveraging Vision Large Language Models

*Jun Yin[1], Jing Zhong[2], Peilin Li[3], Ruolin Pan[4], Pengyu Zeng[5], Miao Zhang[6], Shuai Lu[7]*
*[1,2,5,6,7]Tsinghua University, [3]National University of Singapore,*
*[4]South China University of Technology*
*[1,2,5]{yinj24|zhongj24|zeng-py24}@mails.tsinghua.edu.cn*
*[3]e1351227@u.nus.edu [4]arprl@mail.scut.edu.cn*
*[6,7]{zhangmiao|shuai.lu}@sz.tsinghua.edu.cn*

*Urban cultures and architectural styles vary significantly across cities due to geographical, chronological, historical, and socio-political factors. Understanding these differences is essential for anticipating how cities may evolve in the future. As representative cases of historical continuity and modern innovation in China, Beijing and Shenzhen offer valuable perspectives for exploring the transformation of urban streetscapes.However, conventional approaches to urban cultural studies often rely on expert interpretation and historical documentation, which are difficult to standardize across different contexts. To address this, we propose a multimodal research framework based on vision-language models, enabling automated and scalable analysis of urban streetscape style differences. This approach enhances the objectivity and data-driven nature of urban form research.*
*The contributions of this study are as follows: (1) We construct UrbanDiffBench, a curated dataset of urban streetscapes containing architectural images from different periods and regions. (2) We develop UrbanSense, the first vision-language-model-based framework for urban streetscape analysis, enabling the quantitative generation and comparison of urban style representations. (3) Experimental results show that Over 80% of generated descriptions pass the t-test ($p < 0.05$). High Phi scores (0.912 for cities, 0.833 for periods) from subjective evaluations confirm the method's ability to capture subtle stylistic differences.*
*These results highlight the method's potential to quantify and interpret urban style evolution, offering a scientifically grounded lens for future design.*

## INTRODUCTION

Recent advances in machine learning have led to a wide range of studies in sustainable architecture (Gao et al., 2023; Huang et al., 2025; Li et al., 2019; Zeng et al., 2025; Zhang et al., 2024), energy-aware design (Chen et al., 2024; Jia et al., n.d.; Lu et al., 2024; Lu et al., 2016; Ma et al., 2025), multimodal generation (Sun et al., 2025a; Sun et al., 2025b; Zou et al., 2023), visual enhancement (Yan et al., 2016; Yin et al., n.d.; Zou et al., 2021), architectural performance optimization (Lu et al., 2016; Lu et al., 2024; Ma et al., 2025; Lu et al., 2016;

Yin et al., 2024), and human-AI interaction (Gao et al., 2023; Li et al., 2019; Wang et al., 2024; Yin et al., 2025; Zeng et al., 2025; Zhang et al., 2024). These innovations offer new avenues for studying the evolution of architectural forms.

The evolution of urban architectural styles reflects the trajectories of regional culture, social transformation, and technological advancement. Systematic analysis of urban morphology across different historical periods holds significant academic value — not only for uncovering the mechanisms of stylistic evolution, but also for informing contemporary urban design with historical insights. As two representative cities in China, Beijing and Shenzhen have undergone distinct developmental trajectories: Beijing preserves centuries-old urban scales amid modern planning, while Shenzhen has rapidly emerged over the past four decades from low-density villages prior to the Reform and Opening-up era into a contemporary metropolis. The stylistic transitions shaped by these contrasting contexts merit in-depth investigation.

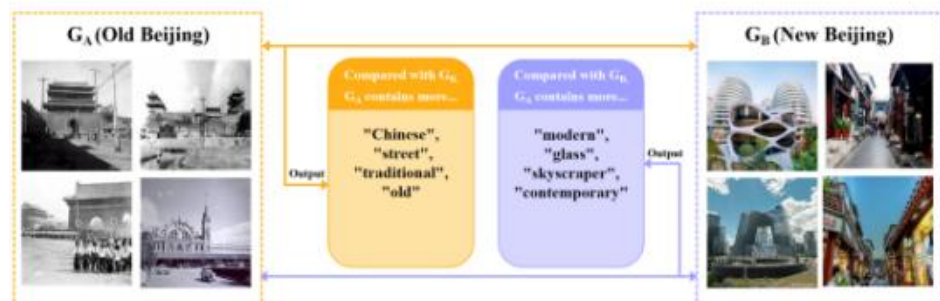

Figure 1
Illustration of set difference captioning for urban streetscape styles. Given image sets $G_A$ and $G_B$, representing historical streetscapes (1950–1970) and modern streetscapes (post-2000) of Beijing respectively, the vision-language model generates natural language descriptions highlighting the most distinctive stylistic features of each period.

While earlier studies provided insights into streetscape styles, traditional methods are hindered by subjectivity, low efficiency, and poor temporal resolution. These limitations have led to growing adoption of AI-driven image – text models for scalable and objective urban style analysis. Recent VLLMs like LLaVA and GPT-4V enhance urban image analysis by combining visual and textual modalities, improving semantic understanding and quantitative reasoning for architectural style recognition.

In this study, we propose an analytical framework, UrbanSense, which leverages state-of-the-art Vision-Language Models (VLLMs) to quantitatively analyze stylistic differences in urban streetscapes across diverse global contexts. This framework integrates deep learning and natural language processing to efficiently extract and generate stylistic descriptions from urban image datasets, enabling comparative analysis across both temporal and geographical dimensions.

UrbanSense offers a novel technical and theoretical approach to architectural style recognition and urban morphology research. By integrating data and interpretability, it links architecture and AI, shifting urban studies from visual observation to semantic analysis.

## RELATED WORK

AI has shifted streetscape analysis from qualitative to data-driven approaches. This section reviews research on style analysis and vision-language models

### Research on urban streetscape style differences

Understanding urban streetscape styles is key to analyzing urban form evolution, yet traditional methods suffer from limited data and subjective labeling (Xu et al., 2023). Recent studies increasingly adopt AI-based techniques, with computer vision—particularly CNNs — enabling large-scale architectural image analysis and reducing reliance on manual annotation (Llamas et al., 2017). However, such approaches often simplify style recognition to surface-level labeling and rarely address temporal stylistic evolution, focusing instead on spatial correlations via embeddings (Han et al., 2022).

### Applications of Vision-Language Models (VLLMs)

Vision-Language Models (VLLMs), particularly CLIP-based variants like UrbanCLIP and UrbanVLP, have been applied to semantically encode satellite imagery and predict socioeconomic factors. UrbanVLP notably introduces a multi-granularity framework for modeling urban functions across regions (Hao et al., 2024). Yet, most VLLM applications remain focused on spatial heterogeneity, with limited exploration of temporal changes in architectural styles or streetscape evolution across historical periods.

## Applications of NLP in visual analysis

Natural Language Processing (NLP), an AI technique for automated language analysis, has seen growing application in urban studies — primarily for extracting semantic insights from sources such as social media opinions and transport-related texts (Cai, 2021).

Despite its potential, NLP in urban visual analysis remains nascent, constrained by manual annotation and prior knowledge for aligning text with architectural elements (Bruschke et al., 2023). While multimodal datasets offer a foundation, unsupervised image-text alignment remains a key unresolved challenge.

## DATASET

To enable accurate streetscape analysis, we constructed the UrbanDiffBench dataset with expert annotations and style-descriptive texts, sourced from platforms like Pinterest®, Behance®, Archdaily®, and Gooood®.

**Beijing:** A total of 230 streetscape images were collected. Among them, 130 depict Beijing during the socialist planned economy period (1950s – 1970s), featuring traditional forms such as siheyuan and hutongs. The remaining 100 images represent post-reform Beijing (2000s onward), illustrating modernization through areas like the CBD near the China World Trade Center.

**Shenzhen:** We collected 230 streetscape images of Shenzhen, including approximately 110 from the pre-reform era — depicting fishing villages and small-town landscapes — and 120 from the post-reform period, reflecting rapid urbanization.

Duplicates were removed and images standardized in size, resolution, and RGB format. Each image was paired with a description, with no filters applied. Data were labeled by city and historical period.

| Old Beijing vs New Beijing | | Old Beijing vs Old Shenzhen | | Old Shenzhen vs New Shenzhen | | New Beijing vs New Shenzhen | |
|---|---|---|---|---|---|---|---|
| This is a black and white photo of people riding bicycles in front of a gate in a city in china | This image shows a futuristic building in a city in china | This is a black and white photo of people riding bicycles in front of a gate in a city in china | The image shows people riding bicycles down a street in a city | A man is walking down a narrow alley in a village in china | The image shows a street in a city with tall buildings and a skyscraper in the background | People crossing a street in a city with tall buildings in the background | This image shows a street in the old part of the city of china |

Figure 2
Image and Text Annotations in the UrbanDiffBench Dataset

## METHODOLOGY

Inspired by GeoCLIP and VisDiff, we developed UrbanSense, a framework for quantifying regional and temporal urban style differences. Built on UrbanDiffBench, it comprises two core modules: UrbanFeature Discoverer for style extraction and UrbanDifference Assessor for variation evaluation.

To address the difficulty of generating descriptions from full datasets $G_A$ and $G_B$, we introduce a two-stage framework. First, the Urban Feature Discoverer extracts representative subsets $A_a$ and $A_b$ to capture style differences. Then, Urban Difference Assessor identifies key features that distinguish the cities' styles..

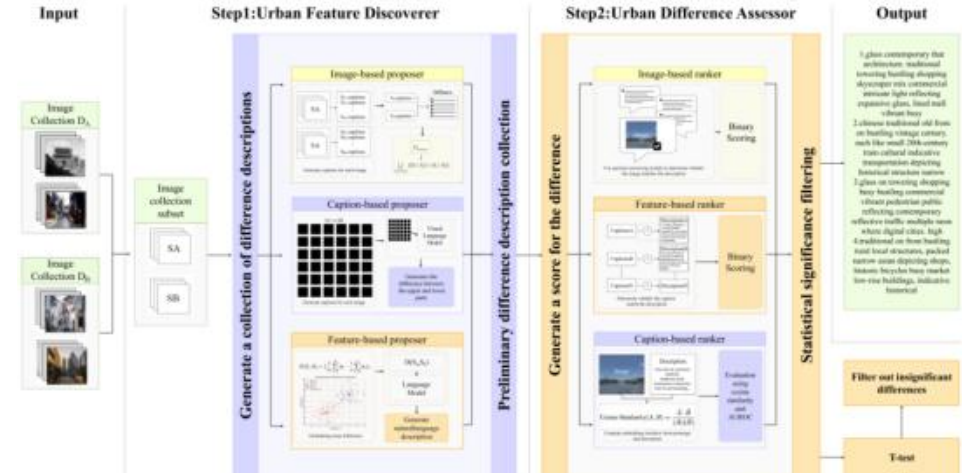

Figure 3
Workflow of the UrbanSense method. The framework consists of two steps: (1) Urban Feature Discoverer extracts descriptive features of urban style; (2) Urban Difference Assessor evaluates and selects the most discriminative features

## Urban Difference Assessor

The Urban Feature Discoverer takes $A_a$ and $A_b$ as input and utilizes a Vision-Language Model (VLM) to output stylistic descriptions with high discriminative power. This process is implemented through the following three approaches:

1. Direct Image-Based Analysis: Two 20-image grids from each subset are input into a VLM (e.g., GPT-4V) to generate descriptive comparisons.
2. Embedding-Based Analysis: Feature embeddings from both subsets are compared,

and their differences are interpreted using a language model (e.g., BLIP-2).

3. Text-Generation-Based Analysis: VLMs generate descriptions for each subset, which are then compared using a language model to identify stylistic differences.

Experiments show that text-generation analysis best captures architectural styles, outperforming visual-only methods. It is thus used as the primary extractor. Multiple samples from subsets $A_a$ and $A_b$ are aggregated and input into the Urban Difference Assessor for deeper analysis.

## Urban Feature Discover

To assess stylistic discrimination, the Urban Difference Assessor computes a discriminative score for each description generated by the Urban Feature Discoverer:

$$D_y = \mathbb{E}_{x \in G_A} s(x, y - \mathbb{E}_{x \in G_B} s(x, y) \qquad (1)$$

Descriptions are ranked by discriminative scores, with a reference value indicating the consistency between each urban image and its generated description. This value is computed using three defined methods:

1. Image-Based: The vision-language model LLaVA-1.5 is queried to assess whether image x matches description y, with its binary output used as the reference score.
2. Text-Generation-Based: BLIP-2 generates a description w for image x, which is then evaluated by Vicuna-1.5 for alignment with y.

Figure 5
Histogram of
Cosine Similarities

3. Feature-Based: CLIP ViT-G/14 is used to compute the embedding vectors of image x and text y, and the cosine similarity between the two vectors is then calculated as follows:

$$s(x, y) = \frac{e_x \cdot e_y}{\|e_x\| \|e_y\|} \qquad (1)$$

Given the continuous nature of the values, AUROC is used to assess each description's ability to distinguish $G_A$ from $G_B$. The feature-based method outperforms others and is adopted as the main evaluation metric, with t-tests applied to retain only statistically significant results ($p < 0.05$).

## EXPERIMENT

We verify the effectiveness of the method through qualitative and quantitative experiments.

## Evaluation of Stylistic Descriptions Effectiveness

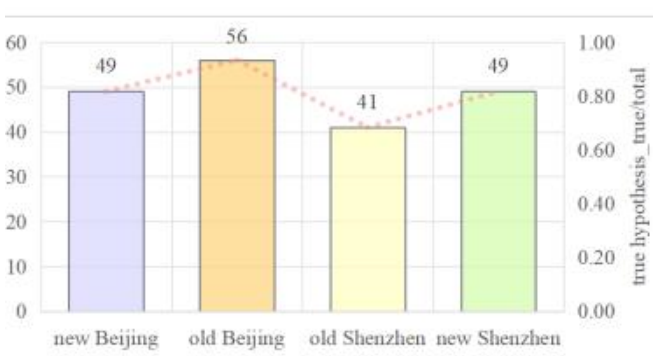

Figure 4
The proportion of descriptions for which the t-test returns a "true"

To assess the effectiveness of the UrbanSense framework, we conducted experiments focusing on the model ' s ability to extract, differentiate, and represent urban stylistic features. The following sections present the evaluation of descriptive accuracy and stylistic distinctions across time and space.

### T-Test and Significance Analysis.

We selected comparison pairs from different regions and periods in the UrbanDiffBench dataset. UrbanSense generated descriptions and computed alignment scores (Score1 and Score2), followed by a t-test to assess the significance of their mean difference.

Four comparison sets produced 240 descriptions. As shown in Figure 4, t-tests show over 80% are statistically significant ($p < 0.05$), confirming the descriptions effectively capture regional and temporal style differences

To quantitatively evaluate the effectiveness of the UrbanSense framework in extracting urban stylistic features, we visualized the group comparison results using three methods: Histogram, Kernel Density Estimation (KDE), and Boxplot.Figure 5 reveals distinct cosine similarity distributions between old and new city pairs. The clear separation of peaks with minimal overlap demonstrates that UrbanSense effectively distinguishes historical from contemporary urban styles, enhancing the model's ability to differentiate stylistic features.The Kernel Density Estimation plot of cosine similarities (Figure 6) shows that UrbanSense identifies both high-similarity peaks and a broad distribution, indicating its ability to capture both concentrated features and diverse stylistic variations.

Figure 6
Kernel Density
Estimation of Cosine
Similarities

Figure 7
Boxplot of Cosine
Similarities

Figure 7 shows that boxplot analysis reveals strong intra-city consistency (e.g., old vs. new Beijing) and marked intercity variation (e.g., new Beijing vs. new Shenzhen), confirming UrbanSense' s ability to capture temporal and spatial style differences.

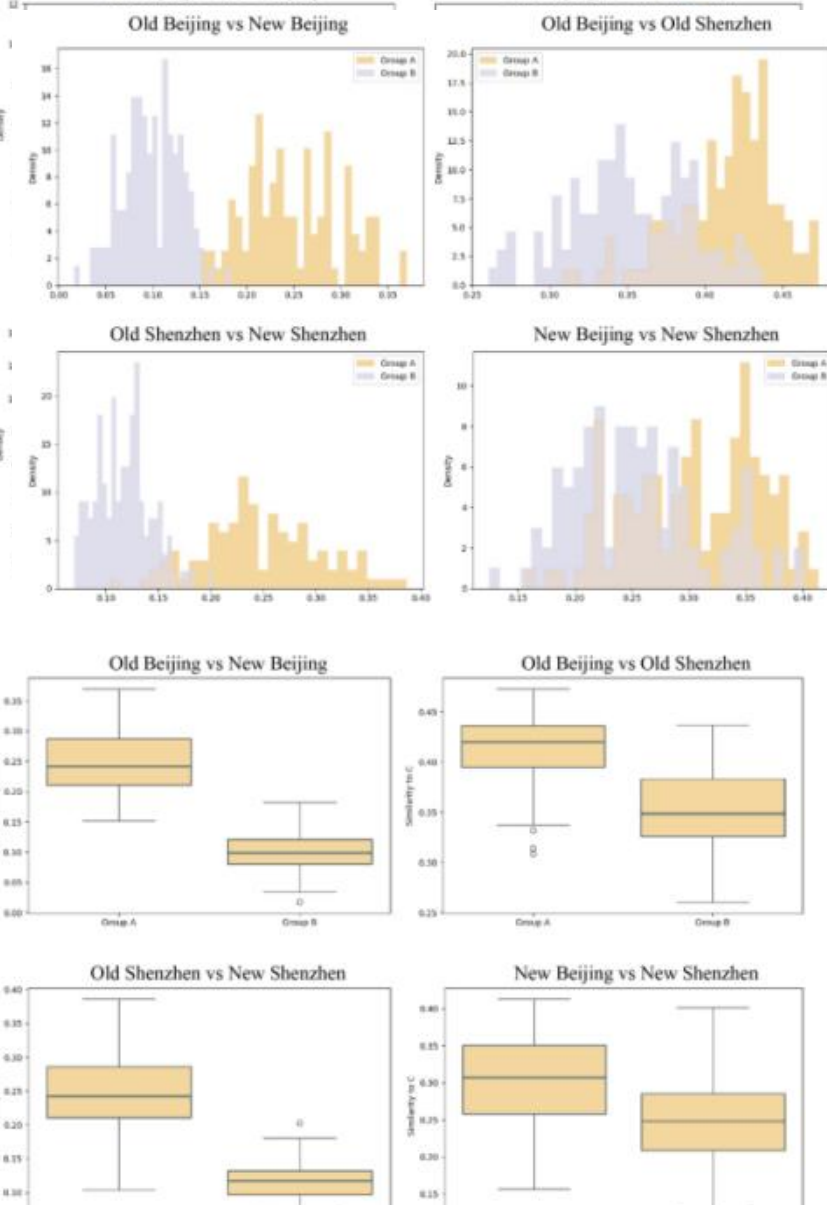

**Interpretation of Urban Style Clustering.**

We applied TF-IDF and PCA to vectorize and reduce 240 descriptions from four urban categories, followed by K-means clustering to examine stylistic distinctiveness across temporal and geographic contexts.

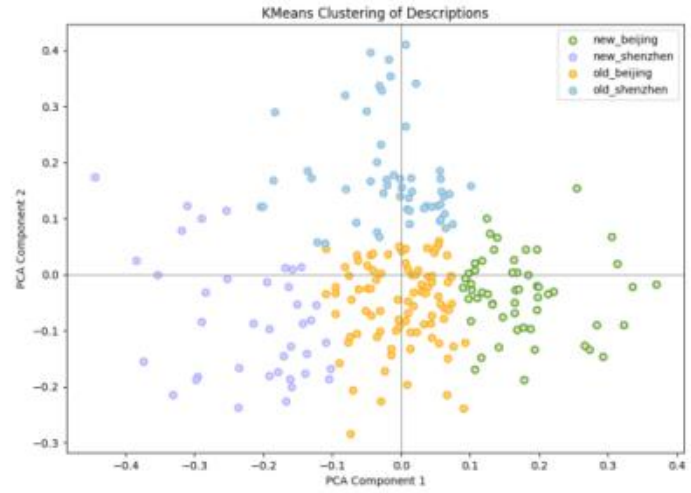

Figure 8
K-Means Clustering Analysis of Stylistic Descriptions

As shown in Figure 8, descriptions from the same category cluster closely, while those from different categories are well-separated, confirming UrbanSense's ability to distinguish heterogeneous stylistic features.The 2D space reveals directional trends in urban style evolution. Shenzhen's old and new clusters are clearly split along the y-axis, indicating a transformative shift, while Beijing's clusters form a continuous distribution, suggesting gradual stylistic evolution.

## Effectiveness of Differential Descriptions in Style Comparison

Building on prior analysis, UrbanSense effectively captures urban style differences. This chapter interprets the morphological evolution of Beijing and Shenzhen through the model's stylistic outputs.

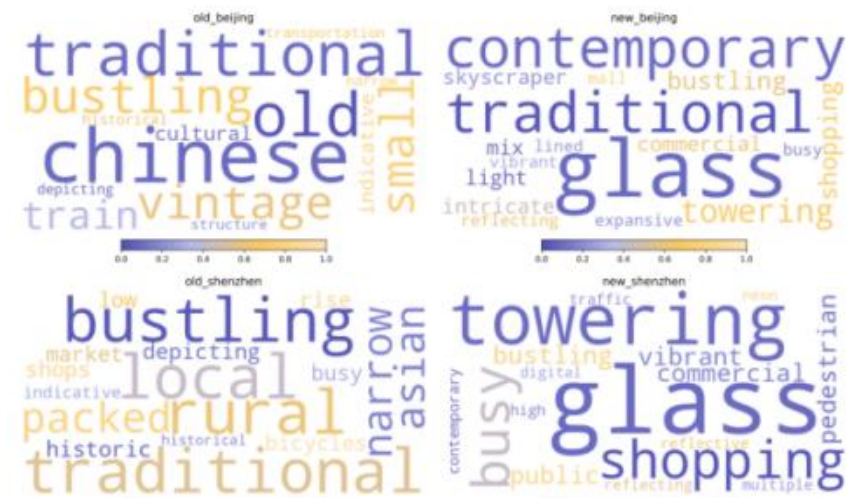

**Extracting Consensus Features of Urban Styles.** BIntegrating architectural history theories, we examine how generated descriptions reflect historical shifts in urban form, validating the interpretability of vision-language methods in urban morphology research.

Using UrbanDiffBench results, we aim to (1) assess the effectiveness of differential descriptions in style comparison, and (2) apply the most discriminative ones as prompts in text-to-image models, with expert validation.

**Extracting Common Stylistic Features of Cities.**
As shown in the word cloud, the model effectively extracts representative visual features that reflect a city's morphological evolution across historical periods.

Taking Beijing as an example, the city’s stylistic evolution is gradual, shifting from traditional elements like “pagoda” and”archway” (1950s–1980s) to modern terms such as “futuristic” and “blending tradition with modernity.” This trajectory aligns with Figure 8’s clustering and the functional–cultural strategy outlined in Urban Memory.In contrast to Beijing’s gradual evolution, Shenzhen shows a sharp stylistic shift — from “low-rise” and “temporary structures” to “towering,” “interchanges,” and “concrete urban setting.” This reflects rapid urbanization, aligning with its spatial reconstruction narrative and Figure 8’s clustering.

The results show that UrbanSense-generated keywords align with widely accepted narratives of urban change while effectively capturing temporal stylistic and spatial differences within a city, demonstrating the model’s strong capability in urban morphological analysis.

Figure 10
Dataset Images with Descriptions Generated by the UrbanSense Model and Their Corresponding Synthesized Visualizations

Figure 11
Dataset Images with Descriptions Generated by the UrbanSense Model

**Identifying Urban Differences Within Similar Stylistic Contexts.**
During the same period, image keywords reveal both shared stylistic traits and divergent urban logics. UrbanSense filters out generic terms, enabling fine-grained interpretation of inter-city differences and generating highly discriminative descriptions.

Keywords like “city walls” and “clock tower” in old Beijing reflect its ceremonial spatial logic, while “small villages” and “narrow streets” in old Shenzhen indicate an organically grown, low-density fabric. Though both cities modernized, new Beijing fuses tradition with vertical urbanism, whereas new Shenzhen emphasizes consumer culture and density — highlighting divergent evolutions rooted in distinct historical and cultural contexts.These results confirm that UrbanSense captures both intra-style nuances and cross-city features within the same era, aligning well with established urban studies and supporting robust multi-city comparisons.

## Expert Evaluation and Validation

To subjectively assess and validate the expressive performance of the UrbanSense model, we invited 10 senior architectural designers and 10 non-architecture-background participants to participate in a structured evaluation.

The first task tested whether observers could identify the urban context of eight images generated by UrbanSense. Participants selected from four categories — Beijing (Planned Economy), Beijing (Contemporary), Shenzhen (Pre-Reform), and Shenzhen (Post-Reform) — to assess the clarity of the model's stylistic representation (Figure 10).

Figure 9
Stylistic Features and Word Clouds of Beijing and Shenzhen in Different Periods

$$Acc_{\text{total}} = \frac{1}{N}\sum_{i=1}^{N}\delta(\hat{c}_i, c_i)\cdot\delta(\hat{p}_i, p_i) \qquad (1)$$

Let $N$ denote the total number of samples; $\delta(a, b)$ denote the Kronecker delta function, which equals 1 when $a = b$, and 0 otherwise. Let $\hat{c}_i$ represent the predicted city of the i-th sample, $\hat{p}_i$ is the predicted period, $c_i$ the ground-truth city, and $p_i$ the ground-truth period for the i-th sample.

Subsequently, Participants completed image-text matching on 8 test sets, each containing 50 images (25 per city) and two corresponding descriptions. A confusion matrix was used to compute the Phi coefficient, measuring consistency in categorical matching. The formula is as follows:

$$\phi = \frac{ad - bc}{\sqrt{(a+b)(c+d)(a+c)(b+d)}} \qquad (1)$$

Figure 11 shows that UrbanSense achieves strong image–text alignment, with accuracy rates of 90.3% (professional) and 85.7% (non-professional). High Phi coefficients (0.912 for city, 0.833 for period) further confirm its semantic clarity and cognitive reliability.

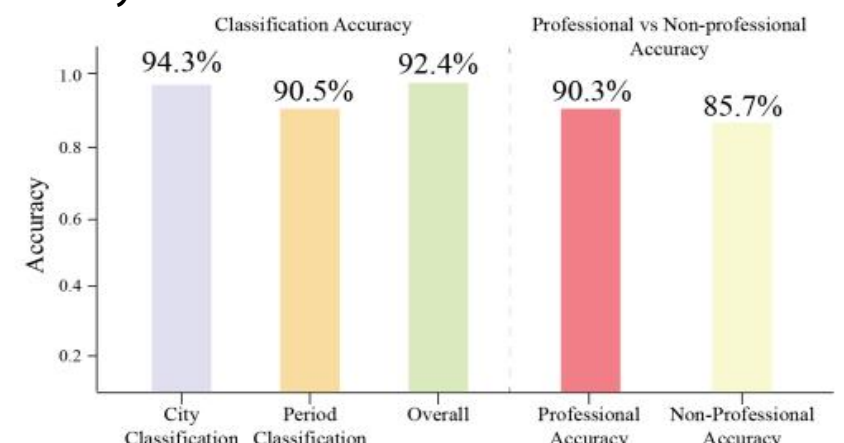

## DISCUSSION

UrbanSense contributes to the automation and objectivity of streetscape interpretation by extracting deep visual features and generating descriptive style-related text. It offers a novel multimodal approach for urban research.

While the framework shows promising potential, challenges remain — such as vague stylistic boundaries and the need for manual text refinement. Future work will focus on improving robustness through larger, more diverse datasets

We acknowledge the current focus on Chinese cities and the need to test generalizability across global contexts, especially where styles are fluid or documentation is sparse. Future work will better situate visual patterns within urban narratives and address dataset bias to ensure cultural fairness.

## CONCLUSION

The UrbanSense framework explores a semantic, quantitative, and systematic pathway for analyzing urban streetscape styles, aiming to provide a technically grounded method for urban morphology research. Experiments validate its effectiveness in style recognition, text generation, and differentiation, with potential applications in design, heritage, and urban renewal.

## AUTHORSHIP INFORMATION

Author Contributions: Conceptualization, J.Y. and P.L.[†]; methodology, J.Y. and P.L.[†]; software, —; validation, J.Z., P.Z., and J.L.; formal analysis, J.Z., P.Z., and M.Z.; investigation, J.Y.; data curation, S.L., M.Z., and J.L.; writing—original draft preparation, J.Y. and P.L.[†]; writing — review and editing, J.Z. and P.L.[†]; visualization, J.Z., L.C., and M.Z.; supervision, P.L.[†]; discussion, J.Z., L.C., and H.D. All authors have read and agreed to the published version of the manuscript.

[*]These authors contributed equally to this work.

[†]correspond author.

## REFERENCE

Xu, H., Sun, H., Wang, L., Yu, X., & Li, T. (2023). 'Urban Architectural Style Recognition and Dataset Construction Method under Deep Learning of street View Images: A Case Study of Wuhan', ISPRS International Journal of Geo - Information, 12(7), 264. Available at: https://doi.org/10.3390/ijgi12070264 (Accessed March 23, 2025)

Llamas, J., M. Lerones, P., Medina, R., Zalama, E., & Gómez-García-Bermejo, J. (2017). Classification

of architectural heritage images using deep learning techniques. Applied Sciences, 7(10), 992.(Accessed March 23, 2025).

Han, Q., Yin, C., Deng, Y., & Liu, P. (2022). 'Towards Classification of Architectural Styles of Chinese Traditional Settlements Using Deep Learning: A Dataset, a New Framework, and Its Interpretability', Remote Sensing, 14(20), 5250. Available at: https://doi.org/10.3390/rs14205250 (Accessed March 23, 2025).

Cai, M. (2021). Natural language processing for urban research: A systematic review. Heliyon, 7.https://doi.org/10.1016/j.heliyon.2021.e06322. (Accessed March 23, 2025)

Vivanco Cepeda, V., Nayak, G. K., & Shah, M. (2023). Geoclip: Clip-inspired alignment between locations and images for effective worldwide geo-localization. Advances in Neural Information Processing Systems, 36, 8690-8701.(Accessed March 23, 2025)

Leng, S., Zhou, Y., Dupty, M. H., Lee, W. S., Joyce, S. C., & Lu, W. (2023). Tell2design: A dataset for language-guided floor plan generation. arXiv preprint arXiv:2311.15941. https://doi.org/10.48550/arXiv.2311.15941.(Accessed March 23, 2025)

Zou, Y., Lou, S., Xia, D., Lun, I. Y., & Yin, J. (2021). Multi-objective building design optimization considering the effects of long-term climate change. Journal of Building Engineering, 44, 102904.

Zhang, M., Shen, Y., Yin, J., Lu, S., & Wang, X. (2024). ADAGENT: Anomaly Detection Agent with Multimodal Large Models in Adverse Environments. IEEE Access.

He, Y., Wang, J., Li, K., Wang, Y., Sun, L., Yin, J., ... & Wang, X. (2024). Enhancing Intent Understanding for Ambiguous Prompt: A Human-Machine Co-Adaption Strategy. Available at SSRN 5119629.

Zhang, M., Yin, J., Zeng, P., Shen, Y., Lu, S., & Wang, X. (2025). TSCnet: A text-driven semantic-level controllable framework for customized low-light image enhancement. Neurocomputing, 129509.

Wang, J., He, Y., Li, K., Li, S., Zhao, L., Yin, J., ... & Wang, X. (2025). MDANet: A multi-stage domain adaptation framework for generalizable low-light image enhancement. Neurocomputing, 129572.

Zeng, P., Jiang, M., Wang, Z., Li, J., Yin, J., & Lu, S. CARD: Cross-modal Agent Framework for Generative and Editable Residential Design. In NeurIPS 2024 Workshop on Open-World Agents.

Yin, J., Gao, W., Li, J., Xu, P., Wu, C., Lin, B., & Lu, S. (2025). Archidiff: Interactive design of 3d architectural forms generated from a single image. Computers in Industry, 168, 104275.

He, Y., Li, S., Li, K., Wang, J., Li, B., Shi, T., ... & Wang, X. (2025). Enhancing Low-Cost Video Editing with Lightweight Adaptors and Temporal-Aware Inversion. arXiv preprint arXiv:2501.04606.

Yin, J., He, Y., Zhang, M., Zeng, P., Wang, T., Lu, S., & Wang, X. (2025). Promptlnet: Region-adaptive aesthetic enhancement via prompt guidance in low-light enhancement net. arXiv preprint arXiv:2503.08276.

Zeng, T., Ma, X., Luo, Y., Yin, J., Ji, Y., & Lu, S. (2025, March). Improving outdoor thermal environmental quality through kinetic canopy empowered by machine learning and control algorithms. In Building Simulation (pp. 1-22). Beijing: Tsinghua University Press.

Zeng, P., Gao, W., Li, J., Yin, J., Chen, J., & Lu, S. (2025). Automated residential layout generation and editing using natural language and images. Automation in Construction, 174, 106133.

Jia, Z., Lu, S., & Yin, J. How real-time energy feedback influences energy-efficiency and aesthetics of architecture design and judgment of architects: two design experiments.

YIN, J., XU, P., GAO, W., ZENG, P., & LU, S. DRAG2BUILD: INTERACTIVE POINT-BASED MANIPULATION OF 3D ARCHITECTURAL POINT CLOUDS GENERATED FROM A SINGLE IM-AGE.

Zeng, P., Hu, G., Zhou, X., Li, S., Liu, P., & Liu, S. (2022). Muformer: A long sequence time-series forecasting model based on modified multi-head attention. Knowledge-Based Systems, 254, 109584.
Sun, H., Xia, B., Chang, Y., & Wang, X. (2025). Generalizing Alignment Paradigm of Text-to-Image Generation with Preferences Through f-Divergence Minimization. Proceedings of the AAAI Conference on Artificial Intelligence, 39(26), 27644-27652. https://doi.org/10.1609/aaai.v39i26.34978
Sun, H., Xia, B., Zhao, Y., Chang, Y., & Wang, X. (2025, April). Identical Human Preference Alignment Paradigm for Text-to-Image Models. In ICASSP 2025-2025 IEEE International Conference on Acoustics, Speech and Signal Processing (ICASSP) (pp. 1-5). IEEE.
Sun, H., Xia, B., Zhao, Y., Chang, Y., & Wang, X. (2025, April). Positive Enhanced Preference Alignment for Text-to-Image Models. In ICASSP 2025-2025 IEEE International Conference on Acoustics, Speech and Signal Processing (ICASSP) (pp. 1-5). IEEE.
Yin, J., Li, A., Xi, W., Yu, W., & Zou, D. (2024, May). Ground-fusion: A low-cost ground slam system robust to corner cases. In 2024 IEEE International Conference on Robotics and Automation (ICRA) (pp. 8603-8609). IEEE.
Zeng, P., Hu, G., Zhou, X., Li, S., & Liu, P. (2023). Seformer: a long sequence time-series forecasting model based on binary position encoding and information transfer regularization. Applied Intelligence, 53(12), 15747-15771.
Zeng, P., Yin, J., Gao, Y., Li, J., Jin, Z., & Lu, S. (2025). Comprehensive and Dedicated Metrics for Evaluating AI-Generated Residential Floor Plans. Buildings, 15(10), 1674. https://doi.org/10.3390/buildings15101674
Li, J., Lu, S., Wang, W., Huang, J., Chen, X., & Wang, J. (2018). Design and Climate-Responsiveness Performance Evaluation of an Integrated Envelope for Modular Prefabricated Buildings. Advances in Materials Science and Engineering, 2018(1), 8082368.
Li, J., Song, Y., Lv, S., & Wang, Q. (2015). Impact evaluation of the indoor environmental performance of animate spaces in buildings. Building and environment, 94, 353-370.
Li, J., Lu, S., & Wang, Q. (2018). Graphical visualisation assist analysis of indoor environmental performance: Impact of atrium spaces on public buildings in cold climates. Indoor and Built Environment, 27(3), 331-347.
Luo Y, Ma X, Li J, Wang C, de Dear R, Lu S. Outdoor space design and its effect on mental work performance in a subtropical climate. Building and Environment. 2025 Feb 15;270:112470.
Lu S, Luo Y, Gao W, Lin B. Supporting early-stage design decisions with building performance optimisation: Findings from a design experiment. Journal of Building Engineering. 2024 Apr 1;82:108298.
Ma, Xintong, Tiancheng Zeng, Miao Zhang, Pengyu Zeng, Borong Lin, and Shuai Lu. "Street microclimate prediction based on Transformer model and street view image in high-density urban areas." Building and Environment 269 (2025): 112490.
Huang, Y., Zeng, T., Jia, M., Yang, J., Xu, W., & Lu, S. (2025). Fusing Transformer and diffusion for high-resolution prediction of daylight illuminance and glare based on sparse ceiling-mounted input. Building and Environment, 267, 112163.
Gao, W., Lu, S., Zhang, X., He, Q., Huang, W., & Lin, B. (2023). Impact of 3D modeling behavior patterns on the creativity of sustainable building design through process mining. Automation in Construction, 150, 104804.
Yu, T., Li, J., Jin, Y., Wu, W., Ma, X., Xu, W., & Lu, S. (2025). Machine learning prediction on spatial and environmental perception and work efficiency using electroencephalography including cross-subject scenarios. Journal of Building Engineering, 99, 111644.

Li, J., Lu, S., Wang, Q., Tian, S., & Jin, Y. (2019). Study of passive adjustment performance of tubular space in subway station building complexes. Applied Sciences, 9(5), 834

Lu, S., Yan, X., Xu, W., Chen, Y., & Liu, J. (2016, June). Improving auditorium designs with rapid feedback by integrating parametric models and acoustic simulation. In Building Simulation (Vol. 9, pp. 235-250). Tsinghua University Press.

Yan, X., Lu, S., & Li, J. (2016). Experimental studies on the rain noise of lightweight roofs: Natural rains vs artificial rains. Applied Acoustics, 106, 63-76.

Lu, S., Yan, X., Li, J., & Xu, W. (2016). The influence of shape design on the acoustic performance of concert halls from the viewpoint of acoustic potential of shapes. Acta Acustica United with Acustica, 102(6), 1027-1044.

Lu, S., Xu, W., Chen, Y., & Yan, X. (2017). An experimental study on the acoustic absorption of sand panels. Applied Acoustics, 116, 238-248.

Zhang, Q., Li, Z., Chen, M., Liu, Y., Chen, Q., & Wang, Y. (2024). Pandora's Box or Aladdin's Lamp: A comprehensive analysis revealing the role of RAG noise in large language models. arXiv preprint arXiv:2408.13533. https://arxiv.org/abs/2408.13533

Wang, J., Liu, C., Zhu, W., He, Y., & Liu, Z. (2024). Beyond examples: High-level automated reasoning paradigm in in-context learning via MCTS. arXiv preprint arXiv:2411.18478. https://arxiv.org/abs/2411.18478

He, Y., Wang, J., Wu, J., Zhu, W., & Liu, Z. (2025). Boosting multimodal reasoning with MCTS-automated structured thinking. arXiv preprint arXiv:2502.02339. https://arxiv.org/abs/2502.02339

Gao, C., Sun, Y., Yang, K., Xie, Y., Liu, J., & Liu, Y. (2025). Thought-augmented policy optimization: Bridging external guidance and internal capabilities. arXiv preprint arXiv:2505.15692. https://arxiv.org/abs/2505.15692

Wang, Y., He, Y., Wang, J., Li, K., Sun, L., Yin, J., ... & Wang, X. (2025). Enhancing intent understanding for ambiguous prompt: A human-machine co-adaption strategy. Neurocomputing, 130415. https://doi.org/10.1016/j.neucom.2024.130415

Yin, J., Zhong, J., Zeng, P., Li, P., Zheng, H., Zhang, M., & Lu, S. (2025). ArchShapeNet: An interpretable 3D-CNN framework for evaluating architectural shapes. *International Journal of Architectural Computing*, 14780771251352965.

Yin, J., Zeng, P., Zhong, J., Li, P., Zhang, M., Luo, R., & Lu, S. (2025). FloorPlan-DeepSeek (FPDS): A multimodal approach to floorplan generation using vector-based next room prediction. *arXiv preprint arXiv:2506.21562*.

Yin, J., Zhong, J., Zeng, P., Li, P., Zhang, M., Luo, R., & Lu, S. (2025). FloorplanMAE: A self-supervised framework for complete floorplan generation from partial inputs. *arXiv preprint arXiv:2506.08363*.

Li, P., Yin, J., Zhong, J., Luo, R., Zeng, P., & Zhang, M. (2025). Segment Any Architectural Facades (SAAF): An automatic segmentation model for building facades, walls and windows based on multimodal semantics guidance. *arXiv preprint arXiv:2506.09071*.